\documentclass{article}
\pdfoutput=1

\usepackage{microtype}
\usepackage{graphicx}
\usepackage{subfigure}
\usepackage{booktabs} % for professional tables

\usepackage{hyperref}

\usepackage[accepted]{icml2024}

\usepackage{amsmath}
\usepackage{amssymb}
\usepackage{mathtools}
\usepackage{amsthm}

\usepackage[capitalize,noabbrev]{cleveref}

\theoremstyle{plain}

\theoremstyle{definition}

\theoremstyle{remark}

\usepackage[textsize=tiny]{todonotes}

\usepackage{stfloats} % To force 'figure*' environment to respect my placement wishes
\usepackage{comment}
\usepackage{enumitem}
\setlist{nosep} % or \setlist{noitemsep} to leave space around whole list

\icmltitlerunning{Capturing Sparks of Abstraction}

\begin{document}

% Where do LLMs stand in terms of performance on cognitive tasks, 
% such as reasoning, navigation, planning, and theory of mind?

%\raggedcolumns
\twocolumn[
\icmltitle{Capturing Sparks of Abstraction for the ARC Challenge}



\icmlsetsymbol{equal}{*}

\begin{icmlauthorlist}
\icmlauthor{Martin Andrews}{rdai}
%\icmlauthor{Sam Witteveen}{rdai}
%\icmlauthor{Firstname2 Lastname2}{equal,yyy,comp}
%\icmlauthor{Firstname3 Lastname3}{comp}
%\icmlauthor{Firstname4 Lastname4}{sch}
%\icmlauthor{Firstname5 Lastname5}{yyy}
%\icmlauthor{Firstname6 Lastname6}{sch,yyy,comp}
%\icmlauthor{Firstname7 Lastname7}{comp}
%\icmlauthor{Firstname8 Lastname8}{sch}
%\icmlauthor{Firstname8 Lastname8}{yyy,comp}
\end{icmlauthorlist}

%\icmlaffiliation{yyy}{Department of XXX, University of YYY, Location, Country}
%\icmlaffiliation{comp}{Company Name, Location, Country}
%\icmlaffiliation{rdai}{Red Dragon AI, Singapore}
\icmlaffiliation{rdai}{Red Dragon AI, Singapore}
%\icmlaffiliation{sch}{School of ZZZ, Institute of WWW, Location, Country}

\icmlcorrespondingauthor{Martin Andrews}{martin@reddragon.ai}
%\icmlcorrespondingauthor{Firstname2 Lastname2}{first2.last2@www.uk}

\icmlkeywords{LLM,Abstraction,ARC Challenge,DSL}

\vskip 0.3in
]



% Next line from https://tex.stackexchange.com/questions/415625/avoiding-hyperref-warning-ignoring-empty-anchor 
\makeatletter\def\Hy@Warning#1{}\makeatother
\printAffiliationsAndNotice{}  % leave blank if no need to mention equal contribution
% NB ::: Commented out %\Notice@String in icml2024.sty to stop the ICML stuff appearing

\begin{abstract}

%Asymptotic performance, apparently missing classes of problems.

%Large Language Model = Gemini-Flash (seems good enough) - benefits from long context / low cost.
%Local LM = model usable within the 2xT4 16Gb constraints (also factor in 10k context for problem description, etc)

Excellent progress has been made recently in solving ARC Challenge problems. 
However, it seems that new techniques may be required to push beyond 60\% accuracy.
Even commercial Large Language Models (LLMs) struggle to `understand' 
%how many of the problems work (given the input and output grids), 
many of the problems (when given the input and output grids), 
which makes discovering solutions by LLM-lead program search somewhat futile.

In this work, LLM `understanding' is attempted from a stronger starting position : 
An LLM is given complete solutions to tasks in code, 
and then asked to explain how the task is being solved at various levels of abstraction.
Specifically, the LLM was given code solutions implemented in {\tt arc-dsl-llm}
(an LLM-legible version of Hodel's {\tt arc-dsl})
%, along with variable values from tracing execution of the code, 
to obtain:
(a) commented code; 
(b) code refactored into reusable functional chunks; 
(c) problem solution steps; and 
(d) high-level problem-solving tactics.  

We demonstrate that `Sparks of Abstraction' can be extracted from the LLM output 
- in a form that could be used in downstream tasks with Local LLMs 
eligible to enter the ARC Prize.

%Want to build a more strategic system 
%GPT4 seems to 'understand' about 50\% of the problems, and fails to code some percentage of them

Both the 
{\tt arc-dsl-llm} DSL framework (with the re-engineered solutions)
and the 
Gemini LLM-generated data (along with the generation code) 
are made Open Source.

%Gemini LLM ARC-training dataset release, along with utilities code and {\tt arc-dsl-llm}.

%Importantly for the ARC Challenge restricted computation environment, the LLM outputs can be (i) used for Local LM fine-tuning; and (ii) made %accessible via a simple RAG system for injection into live prompts on test-set problems.

\end{abstract}

\section{Introduction}
\label{introduction}

The ARC dataset introduced in \citet{chollet2019measureintelligence} has remained 
strikingly resistant to the scale-up in compute power 
that has lead to the quick obsolescence of many other benchmarks.
Interestingly, both commercial LLMs (such as GPT-4o)
and systems that are eligible to enter the ARC Prize competition have arrived 
at approximately the same scores on the challenge - 
potentially suggesting that current approaches are unable to grapple with entire classes of problems.
%Asymptotic performance, apparently missing classes of problems.
%
%
One possibility is that there are elements of abstraction and strategic thinking that are missing.

\begin{figure}
  \centering
  \includegraphics[width=0.9\linewidth]{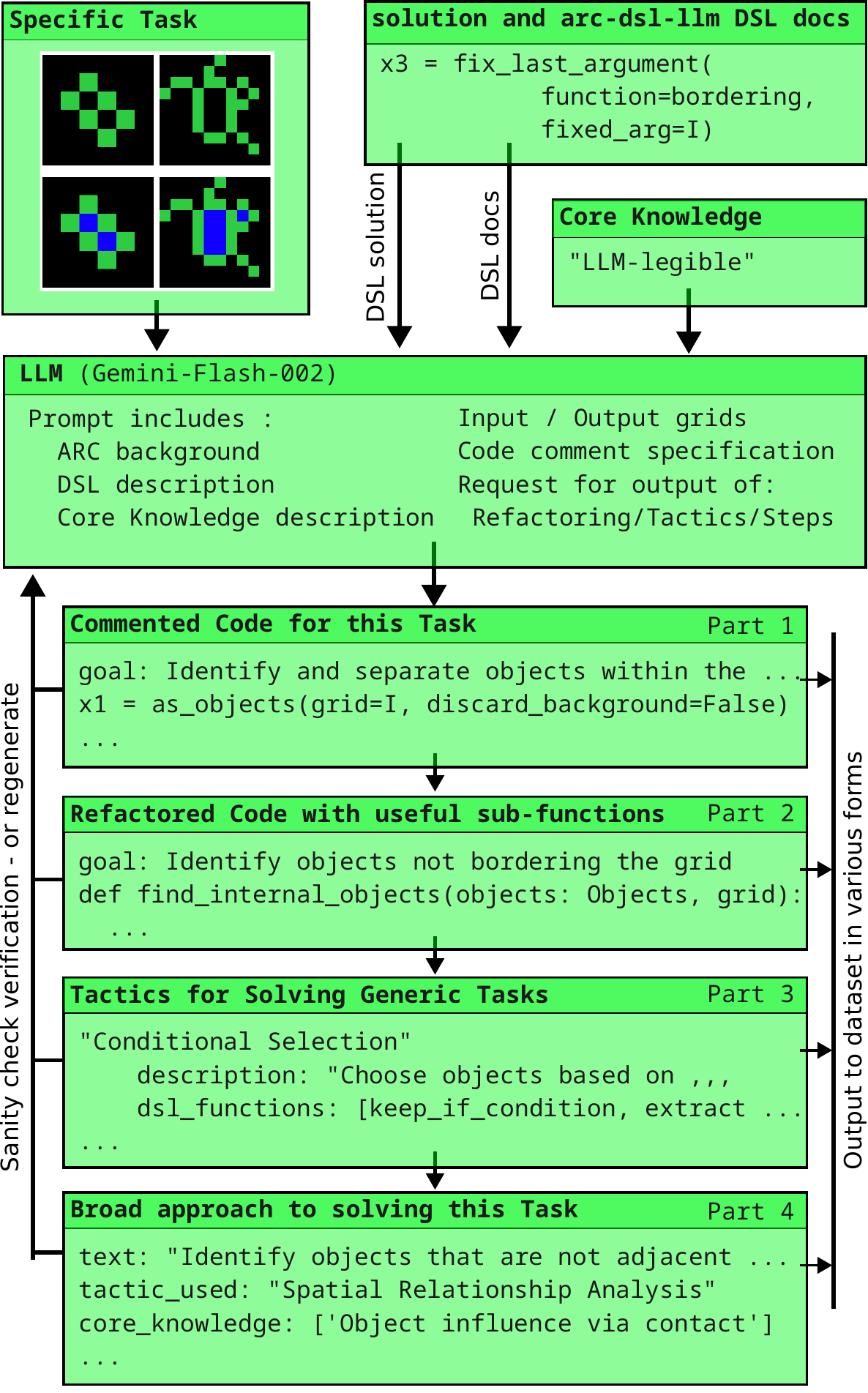}
  \caption{Information flow from Task 00d62c1b to Dataset}
  \label{InfoFlowDiagram}
\end{figure}

In this work, we use Gemini-Flash 
%(a mid-tier commercial LLM, capable of using a long context window, while being an order of magnitude cheaper than frontier models) 
to `reason' about known-good code solutions to ARC training set problems.
These solutions are expressed in {\tt arc-dsl-llm}, 
which is an adaptation of {\tt arc-dsl} released in \citet{hodel-repo-arc-dsl}, 
re-engineered to be more `LLM-legible'.
%The re-engineering effort was undertaken to make the DSL functions, 
%and the solutions to the training-set problems previously implemented, 
%more LLM-legible - since in its original form, the DSL was not particularly human-readable.

%Want to build a more strategic system
%GPT4 seems to 'understand' about 50\% of the problems, and fails to code some percentage of them

%Large Language Model = Gemini-Flash (seems good enough) - benefits from long context / low cost.
%Local LM = model useable within the 2xT4 16Gb constraints (also factor in 10k context for problem description, etc)

The main idea behind our approach is that LLMs (due to their training) 
have some facility in dealing with code - and this `understanding'
can be captured for processing further downstream.
See Figure \ref{InfoFlowDiagram} for an outline of the process.
%
%As illustrated 
Overall, it appears that ``Sparks of Abstraction'' are present in the LLM outputs generated, 
i.e. the LLM has `understood the big picture' for the problem, 
even if the code (expressed via DSL) approached the problem in an obscure way.

Importantly for the ARC Challenge restricted computation environment, 
the LLM outputs can be 
(i) used for Local LM fine-tuning; and 
(ii) made accessible via a simple RAG system for injection into live prompts on test-set problems.

%which can then be output for processing for Local LMs) 
%clearly has more applications than just the current dataset.

%We used the LLM to leverage the code solutions from `arc-dsl` (enhanced as a more LLM-legible `arc-dsl-llm`)%
%, along with variable values from tracing execution of the code, 
%to obtain:
%(a) commented code; 
%(b) code refactored into reusable functional chunks; 
%(c) problem solution steps; and 
%(d) generally applicable high-level tactics.  

%Both `arc-dsl-llm` and the LLM-generated data (along with the generation code) are released online.
%Gemini LLM ARC-training dataset release, along with utilities code and `arc-dsl-llm`.

\begin{comment}

\begin{figure}
  \centering
{\footnotesize
%\begin{alltt}
\begin{verbatim}
# (1) Statement of original problem
def proof(
      clue="arrived with an artist, \
              to get optical device", 
      pattern="6", 
      answer="CAMERA"):  # Provided
  """
  # (2) Hypothesised by local LM
  definition: arrived with an artist, \
                to get {optical device}
  wordplay: CAME (arrived) + \
    RA (artist, from RA = Royal Academy)
  """
  # (3) Continuation generated by LLM 
  assert is_synonym("arrived", "CAME")
  assert is_abbreviation("artist", "RA")
  assert "CAME" + "RA" == "CAMERA"
  assert is_synonym(
     "optical device", "CAMERA",  
     pattern="6")
proof()   # Triggers proof verification
\end{verbatim}
%\end{alltt}
}
\caption{From problem statement to LLM formalisation}
\label{WordplayFormalisation}
\end{figure}

\end{comment}

\subsection{Contributions}
\label{Contributions}

The following are the main contributions of this work\footnote{
NB: No {\it serious} entry into the Kaggle ARC Prize competition was made, 
since the current goal is to solve the `missing class' problem, rather than battling over the easier tasks
}%
:

\begin{itemize}  % bulleted list
\item {\bf LLM-legible ARC DSL} - 
{\tt arc-dsl-llm} is a version of {\tt arc-dsl} designed to be more readable, 
with additional fixes for correctness and type-safety.
We release both the enhanced DSL code, and solutions for all 400 ARC training tasks,
at \url{https://github.com/mdda/arc-dsl-llm}
% Dataset released as a concrete contribution.
\item {\bf Dataset release} - 
Outputs from Gemini-Flash-002 for the 377 ARC training tasks which have passed the sanity checking process,
along with the generation code and DSL manipulation utilities,
are made available at \url{https://github.com/mdda/LLM-abstraction-for-ARC}
\end{itemize}

\section{Related Work}
\label{RelatedWork}

%\subsection{Cryptic Crosswords}
%\label{Cryptic Crosswords}

\subsection{Core Knowledge}
\label{Core Knowledge}

An often overlooked resource for the abstractions used in the ARC challenge tasks is the original ARC paper 
\citep{chollet2019measureintelligence}.  
The Core Knowledge described there, building on \citet{spelke2007core}, 
can be seen as the obvious basis for computer approaches at solving the tasks - 
there is no need to reach for more complex abstractions 
(since they are unlikely to be accessible to most humans).
%indeed, the ARC tasks are clearly `General' (in the AGI sense)
%
%* [On the Measure of Intelligence](https://arxiv.org/abs/1911.01547)
%  - [ARC repo - fixed recently](https://github.com/fchollet/ARC/)
%
This idea was further explored in \citet{moskvichev2023conceptarcbenchmarkevaluatingunderstanding}.
%$* [ConceptARC Benchmark: Evaluating Understanding and Generalization in the ARC Domain](https://arxiv.org/abs/2305.07141)
%  + Melanie Mitchell
%  + [Code Repo](https://github.com/victorvikram/ConceptARC)
%  + Has human solving rate data

During preliminary work, the LARC dataset introduced in \citet{acquaviva2023communicatingnaturalprogramshumans} was explored.
However, it was found (from the raw data) that humans communicating about the task to be performed were surprisingly bad narrators - 
and the key results of the LARC paper focused on the best performance on each task, rather than overall performance.

%* Language ARC : [Communicating Natural Programs to Humans and Machines](https://arxiv.org/abs/2106.07824)
%  + [Repo](https://github.com/samacqua/LARC)
%    - [Explorer interface - with language descriptions](https://samacqua.github.io/LARC/explore)
%    - has all the descriptions for the 400 (training) tasks
%  + 88\% of ARC tasks can be communicated simply from human to human
%    - 1) natural programs use a much wider range of concepts compared to a typical DSL; 
%    - 2) natural programs contain clarifications and validations in greater quantity than directly executable procedures.
%  + In Appendix page 15 : UI for Describer + Builder.  Describer doesn't see test input, builded doesn't see train pairs.
%  + In Appendix page 16 : 17 classes of kinds of descriptions
%  + In Appendix page 18 : 103 primitives in their DSL (derived from human descriptions)
%  + [language guided program synthesis results for LARC](https://github.com/theosech/ec/tree/language-guided_program_synthesis_for_larc)
%  + [GPT-4 traces](https://github.com/evanthebouncy/larc_gpt4/tree/master/gpt4_no_vision/larc_csv)

\subsection{Domain Specific Language (DSL) for ARC}

One of the foundations of this work is the extraordinary contribution of \citet{hodel-repo-arc-dsl},
which included both the {\tt arc-dsl} implementation, and the solution of the 400 training set ARC tasks written using the DSL.

While our methods could be extended to include code written in arbitrary Python, 
it made sense to use {\tt arc-dsl} as a starting point - 
the value of the known-good solutions outweighing some of the implementation quirks described in Section \ref{arc-dsl-llm-quirks}.

%Hodel
%  + Michael Hodel / re-arc [Reverse Engineering the Abstraction and Reasoning Corpus](https://github.com/michaelhodel/re-arc)
%  + [Addressing the Abstraction and Reasoning Corpus via Procedural Example Generation](https://arxiv.org/abs/2404.07353)
%* [ARC-DSL](https://github.com/michaelhodel/arc-dsl)
%  + https://github.com/michaelhodel/arc-dsl/blob/main/solvers.py

\subsection{LLM Capabilities}

The use of LLMs to solve ARC tasks attracted an initial wave of optimism : 
%(\citet{tan2023largelanguagemodelllm}, \citet{wang2024hypothesissearchinductivereasoning}, \citet{greenblatt2024arcagi})
%orer
\citet{tan2023largelanguagemodelllm,wang2024hypothesissearchinductivereasoning,greenblatt2024arcagi}.

However, there is now mounting evidence that merely scaling the number of samples is unlikely to be an effective way to solve tasks
that involve more abstraction or compositional reasoning \citet{brown2024largelanguagemonkeysscaling}.

\subsection{Problem tactics}

The ability of LLMs to reason was explore in \citet{lee2024reasoningabilitieslargelanguage}, which concluded 
``although current LLMs exhibit outstanding performance, 
they lack logical coherence, compositionality, and productivity in their processes, 
suggesting that they are closer to probabilistic
mimicry rather than possessing autonomous reasoning abilities''.
This suggests that a strategy other than generation from scratch
is required for extracting `higher-level' thinking from LLMs.  
Thus, in this work we start from the basis of extracting these `higher-level' ideas from code created by humans.
%
%\citet{lee2024reasoningabilitieslargelanguage}
%  + [Reasoning Abilities of Large Language Models: In-Depth Analysis on the Abstraction and Reasoning Corpus](https://arxiv.org/abs/2403.11793)
%Experimental results confirm that while large language models possess weak inference abilities, 
%they still lack in terms of logical coherence, compositionality, and productivity. 
%Our experiments highlight the reasoning capabilities of LLMs, proposing development paths for achieving human-level reasoning.

Drawing from the Self-Discover concept of \citet{zhou2024selfdiscoverlargelanguagemodels}, 
this paper aims to extract useful `high-level' reasoning structures for ARC tasks 
(in addition to refactored code examples and Core Knowledge observations).

%\citet{zhou2024selfdiscoverlargelanguagemodels}
%* [Self-Discover: Large Language Models Self-Compose Reasoning Structures](https://arxiv.org/abs/2402.03620)
%  + DeepMind paper 
%  + Framework for LLMs to self-discover the task-intrinsic reasoning structures 
%    - to tackle complex reasoning problems that are challenging for typical prompting methods

%* [ExeDec: Execution Decomposition for Compositional Generalization in Neural Program Synthesis](https://arxiv.org/abs/2307.13883)
%  + DeepMind : https://deepmind.google/discover/blog/google-deepmind-at-iclr-2024/
%  + decomposition-based synthesis strategy that predicts execution subgoals 
%    to solve problems step-by-step informed by program execution at each step
%  + benchmarks demonstrate that LLMs struggle to compositionally generalize 
%    when asked to do programming-by-example in a few-shot setting, 
%    but an ExeDec-style prompting approach can improve the generalization ability and overall performance.

\subsection{Code Generation}

As observed in \citet{greenblatt2024arcagi}, LLMs are limited in their capability of creating new code.
This will clearly also hinder program-search using a Local LM 
(as would be used in the compute-limited Kaggle environment for the ARC Prize competition).

Therefore, if we want to attempt approaches such as 
code denoising \citep{kapur2024diffusionsyntaxtreesprogram},
RL in program space \citep{butt2024codeitselfimprovinglanguagemodels}, 
or DreamCoder \citep{ellis2020dreamcodergrowinggeneralizableinterpretable},
the system would likely benefit from having well annotated code examples,
along with high-level goals to act as in-context prompts.
This motivates this work's attempt at exploring what is possible to extract from LLMs 
that can only be accessed `outside the Kaggle box'.

However, LLM interactions are not the main objective :
All the methods here have been developed such that the extracted data can be hosted `inside the Kaggle box' 
(through RAG, etc).
This is also the rationale for this work's title: ``Capturing the Sparks of Abstraction...''

\begin{comment}

\footnote{
Cracking the Cryptic (17-May-2024) \\
\url{https://youtu.be/vudt7LlUX00?t=124}
}.

\end{comment}

\newpage
\section{Methods}
\label{Methods}

% ## Priors available to the LLM

In order to get the best results from an LLM, it is essential to play to its strengths 
(which are, after all, based on reading a huge quantity of text and code from the internet).
This Section illustrates how we address (and potentially harness) the {\it priors} that the LLM is likely to have.

Firstly, since LLMs are trained on text that is largely designed to be human readable,
text that is less human-readable is likely to be less familiar,
and thus more difficult to extrapolate from (i.e. it is also less LLM-legible).
So, we assume here that human-legibility can be used as a simple proxy for LLM-legibility
(and also note that without training via Reinforcement Learning, 
LLMs do not have any insight into what would increase LLM-legibility).

\subsection{Core Knowledge}

The original textual description of the Core Knowledge from \citet{chollet2019measureintelligence} 
was reworked until the LLM was satisfied that it was clear and interpretable.
The full text is given in Appendix \ref{appendix-core-knowledge}.
%* Dataset release includes an LLM-refined version of the Core Knowledge (to increase, indirectly, LLM-legibility)

\subsection{LLM-legible DSL}
\label{arc-dsl-llm-quirks}

Preliminary work with the {\tt arc-dsl} of \citet{hodel-repo-arc-dsl} suggested that
not only was the DSL code difficult to read by human coders, 
but also that there was a significant risk that an LLM would have difficulty.
For instance, some of the DSL functions were given names that contradict common Python usage - 
an example being {\tt fork()} used to denote a function applied to two different function applications :
it was renamed to {\tt \verb|combine_two_function_results|()}.
Following that, a large number of other similar changes were made.
The solutions have also been re-written, 
and have been validated against the known test solutions in the ARC training set.

%The re-engineering effort was undertaken to make the DSL functions, 
%and the solutions to the training-set problems previously implemented, 
%more LLM-legible - since in its original form, the DSL was not particularly human-readable.

A comparison between the two DSLs is given in Figure \ref{TwoDSLComparison},
and further details about {\tt arc-dsl-llm} are given in Appendix \ref{appendix-arc-dsl-llm}.
Note that the new solver function {\tt\verb|solver_virtual(I)|} returns a dictionary of all the useful intermediate values, 
so that these can be used in subsequent analysis.
%, 
%and the code is available at \url{https://github.com/mdda/arc-dsl-llm} under an Open Source license.

\begin{figure*}[!t]
{\scriptsize%
\begin{minipage}[t]{0.33\linewidth}%
Original {\tt arc-dsl}
\begin{verbatim}
def solve_00d62c1b(I):
    x1 = objects(I, T, F, F)
    x2 = colorfilter(x1, ZERO)
    x3 = rbind(bordering, I)
    x4 = compose(flip, x3)
    x5 = mfilter(x2, x4)
    O = fill(I, FOUR, x5)
    return O
\end{verbatim}%
\end{minipage}%
\begin{minipage}[t]{0.67\linewidth}%
LLM-legible {\tt arc-dsl-llm}
\begin{verbatim}
def solver_virtual(I):
  x1 = as_objects(grid=I, discard_background=False)
  x2 = color_filter(objs=x1, color=COLOR_ZERO)
  x3 = fix_last_argument(function=bordering, fixed_arg=I)
  x4 = compose(outer=logical_not, inner=x3)
  x5 = keep_if_condition_and_flatten(container=x2, condition=x4)
  O = fill(grid=I, color=COLOR_FOUR, patch=x5)
  return dict(I=I,x1=x1,x2=x2,x5=x5,O=O)
\end{verbatim}%
\end{minipage}%
}
\caption{Code comparison between original, and LLM-legible versions of the same code}%
%\caption{Code comparison between Comparing {\tt arc-dsl} (LHS) to {\tt arc-dsl-llm} (RHS)}
\label{TwoDSLComparison}
\end{figure*}

\subsection{Actual coded solutions}

%In addition,  
%to the starting point for {\tt arc-dsl-llm}, 
This work aims to capitalise on the solutions to the 400 ARC training set problems provided by \citet{hodel-repo-arc-dsl}.
%use of two key contributions from : 
%(i) construction of a useful DSL; and (ii) 
%
Of course, since complete solutions to 400 problems are available, the LLM does not have to start with blind search,
it could be train (for instance) on completions from any point, or on a de-noising task.
In addition, each code solution can be assumed to be meaningful (i.e. each line was written with intentionality),
which makes the goal of explaining what each line is doing achievable.

\subsection{Code comments}

It is common practice for programmers to write code comments to explain what the code is supposed to do to the next viewer.
Thus, since LLMs are trained on commented code, and appear to have some skill at writing comments,
it is reasonable to hope that an LLM might recognise some `intentionality' from valid code.
Clearly, there may be elements of abstraction being used here (depending on the size of the code block being described).

\subsection{Language Model targets}

Throughout this work% 
%\footnote{
%NB: no serious entry into the Kaggle ARC Prize competition was made, 
%since the current goal is to solve the `missing piece' problem, rather than the easier tasks
%}
,
two classes of Language Model have been treated as targets for learning about and making use of abstraction:
\begin{itemize}
\item {\bf Large Language Model} - the Gemini-Flash-002 model was chosen (after also testing Gemini-Pro), 
since although it is not a frontier commercial LLM, 
it is capable of using a long context window, 
while being an order of magnitude cheaper than frontier models
\item {\bf Local LMs} - models that are usable within the constraints of the Kaggle competition run-time container 
(i.e. 2xT4 with 16Gb GPU RAM each, 
where we must also factor in approximately 10k tokens of context for the problem description, etc)
\end{itemize}

The dataset released by this work consists of outputs from the Gemini-Flash LLM that have gone through some sanity checks:
(a) they have the required number of `Parts' output; 
(b) the parts are valid Python/YAML as required; 
plus other factors that are given per-Part in Section \ref{Results}.

\begin{comment}

\end{comment}

\section{Results}
\label{Results}

For the outputs shown in the Figures in this Section, 
we use the illustrative ARC task of Figure \ref{InfoFlowDiagram}.
This task was chosen for reasons of brevity (rather than being cherry-picked).
Note though, that this example also illustrates the way in which many of the solutions 
from \citet{hodel-repo-arc-dsl} are coded in a somewhat unorthodox style
(i.e. the code has a very functional style versus a more natural Pythonic approach).

Note, though, that each ARC task is analysed by the LLM independently, 
and the LLM used (Gemini-Flash-002, a pinned version) was not fine-tuned on any data.
Thus, the outputs shown here have not brought information from other ARC problems to bear - 
only the Core Knowledge and the DSL functional descriptions 
(plus the priors that come from the implementations of the solution code itself).

\subsection{Part 1 : Code Commenting}

\begin{figure*}[ht]
{\scriptsize%
\begin{verbatim}
def solver_virtual(I):
  # Input: I (Grid), the input grid.
  # Goal: Identify and separate objects within the input grid based on color and connectivity.
  # Output: x1 (Objects), a set of objects identified in the input grid.
  # Core Knowledge: Object cohesion (parsing grids, identifying distinct objects based on spatial contiguity)
  x1 = as_objects(grid=I, discard_background=False)

  # Input: x1 (Objects), a set of objects identified in the input grid.
  # Goal: Filter the objects to keep only those that are black.
  # Output: x2 (Objects), a subset of x1 containing only black objects.
  # Core Knowledge: Object cohesion (filtering objects based on color)
  x2 = color_filter(objs=x1, color=BLACK)

  # Input: x2 (Objects), a set of black objects; I (Grid), the input grid.
  # Goal: Identify black objects that are not bordering the grid.   \
          This effectively selects the internal black objects.
  # Output: x5 (FrozenSet), a set of indices representing the locations of the internal black objects.
  # Core Knowledge: Object influence via contact (bordering), Basic Geometry and Topology priors (relationships)
  x3 = fix_last_argument(function=bordering, fixed_arg=I)
  x4 = compose(outer=logical_not, inner=x3)
  x5 = keep_if_condition_and_flatten(container=x2, condition=x4)

  # Input: I (Grid), the input grid; x5 (FrozenSet), indices of internal black objects; color=BLUE.
  # Goal: Fill the locations specified by x5 in the input grid with blue color.  \
          This is the final step of transforming the input into the output grid.
  # Output: O (Grid), the output grid after filling the internal black object locations with blue.
  # Core Knowledge: Object manipulation (painting/filling), Compositionality
  O = fill(grid=I, color=BLUE, patch=x5)
  return O
\end{verbatim}%
}
\caption{Line-by-line commentary generated by the LLM in the specified format}%
\label{Part1_CodeCommentary}
\end{figure*}

The ability of the LLM to add straight-forward comments to the code from Figure \ref{TwoDSLComparison} 
is illustrated in Figure \ref{Part1_CodeCommentary}.  

The prompt for the LLM (given in full in Appendix \ref{Appendix_Part1}) includes instructions for:
\begin{itemize}
\item {\bf General ARC Rubric} - see Appendix \ref{appendix-gemini-prompt}
\item {\bf Core Knowledge} - see Appendix \ref{appendix-core-knowledge}
\item {\bf DSL documentation} - see the {\tt arc-dsl-llm} code release
\item {\bf Problem solution} - expressed in {\tt arc-dsl-llm} - see Figure \ref{TwoDSLComparison}
\item {\bf Input / Output grids} - this is in a one-token per pixel format, which works fine despite not being valid Python
\item {\bf Optional : Interim variable values} - we can also extract these, since we have valid code and inputs
\item {\bf Instruction about output formats} - centers on comment style for Part 1
\end{itemize}

Figure \ref{Part1_CodeCommentary} shows that the LLM has picked up on some of the finer details of this problem's solution.
While the obvious human interpretation of one of the sub-goals here is ``Choose the internal areas'',
the DSL implementation reads ``Choose all objects that are black which don't touch the edges of the grid''.
The LLM has `realised' this with the third {\tt Goal} comment : ``This effectively selects the internal black objects''.

\newpage
\subsection{Part 2 : Code Refactoring}

\begin{figure*}[ht]
{\scriptsize%
\begin{verbatim}
def identify_objects(grid: Grid) -> Objects:
  # Input: grid (Grid), the input grid.
  # Goal: Identify and separate objects within the input grid based on color and connectivity.
  # Output: objects (Objects), a set of objects identified in the input grid.
  # Core Knowledge: Object cohesion (parsing grids, identifying distinct objects based on spatial contiguity)
  return as_objects(grid=grid, discard_background=False)
    
def filter_objects_by_color(objects: Objects, color: Color) -> Objects:
  # Input: objects (Objects), a set of objects; color (Color), the color to filter by.
  # Goal: Filter the set of objects to keep only those of the specified color.
  # Output: filtered_objects (Objects), a subset of the input objects containing only objects of the specified color.
  # Core Knowledge: Object cohesion (filtering objects based on color)
  return color_filter(objs=objects, color=color)

def find_internal_objects(objects: Objects, grid: Grid) -> FrozenSet:
  # Input: objects (Objects), a set of objects; grid (Grid), the input grid.
  # Goal: Identify objects that are not bordering the grid.
  # Output: internal_objects (FrozenSet), a set of indices representing the locations of the internal objects.
  # Core Knowledge: Object influence via contact (bordering), Basic Geometry and Topology priors (relationships).
  is_internal = compose(outer=logical_not, inner=fix_last_argument(function=bordering, fixed_arg=grid))
  return keep_if_condition_and_flatten(container=objects, condition=is_internal)

def fill_grid(grid: Grid, color: Color, patch: FrozenSet) -> Grid:
  # Input: grid (Grid), the input grid; color (Color), the color to fill with; patch (FrozenSet), indices to fill.
  # Goal: Fill the specified indices in the grid with the given color.
  # Output: filled_grid (Grid), the grid after filling the specified indices.
  # Core Knowledge: Object manipulation (painting/filling)
  return fill(grid=grid, color=color, patch=patch)


def solver_virtual_chunked(I):
  # Input: I (Grid), the input grid.
  # Goal: Process the input grid to produce the output grid according to the specified transformation rules.
  # Output: O (Grid), the transformed output grid.
  # Core Knowledge: Compositionality (combining multiple steps to achieve the overall transformation)

  x1 = identify_objects(I)
  x2 = filter_objects_by_color(x1, BLACK)
  x5 = find_internal_objects(x2, I)
  O = fill_grid(I, BLUE, x5)
  return O
\end{verbatim}%
}
\caption{Refactoring generated by the LLM in the specified format}%
\label{Part2_Refactoring}
\end{figure*}

The ability of the LLM to perform a light refactoring of the code is illustrated in Figure \ref{Part2_Refactoring}.
The aim of this Part is to implement chunking (that, for instance DreamCoder \citep{ellis2020dreamcodergrowinggeneralizableinterpretable} might perform at considerable cost) `outside the box' - 
i.e. the found sub-functions can be exported into a Kaggle container 
(the consistent commenting style makes them RAG-able).

The prompt (given in full in Appendix \ref{Appendix_Part2}) extends Part 1:
\begin{itemize}
\item {\bf What makes a good sub-function } - which is really a matter of taste...
\item {\bf Free to rename variables in sub-functions } - to see whether the LLM can provide its own self-commenting
\item {\bf Main function required name } - the main entry point must be {\tt\verb|solver_virtual_chunked(I)|}
\item {\bf Main function retains original variable names } - so that each code sample can be indexed against the others
\item {\bf Sub-functions should not call each other } - so that they are independently useful `inside the box'
\end{itemize}

A number of sanity-check are applied to the refactored code output, which accounts for most of the missing entries in the dataset.
These are not necessarily the most difficult/longest code solutions, the issue that the LLM seems to have is that
some refactoring `ideas' are so compelling that it cannot resist trying to apply them.  The current checks include:
\begin{itemize}
\item {\bf Sub-functions cannot call each other } - this can be checked to some degree by analysing the code, 
though sometimes there may be in-line functional-synthesis-style calls that are difficult to pick up
\item {\bf The main function must have consistent variable names } - this can be difficult for LLMs to understand, 
since the prompts encourage variable renaming within new sub-functions 
(this may desirable for when they are later deployed in a RAG context).
This consistency requirement also makes the output code less Pythonic (to the detriment of the LLM)
\end{itemize}

%* DreamCoder not required
%* Even if chunking is not perfect, even small Local code LMs can manipulate reliably
%  + `def find_object_with_most_red(objects: Objects) -> Object`

%`TODO`: `code refactoring example` (chunking is not very sophisticated, though)

Overall the refactoring demonstrated in Figure \ref{Part2_Refactoring} 
is not particularly noteworthy (many of the sub-functions suggested are one-liners, despite the prompts given).
However, one interesting observation is that {\tt\verb|find_internal_objects|} has been factored out - 
further emphasising that the LLM has `understood' the goals of this (rather convoluted) method.

\newpage
\subsection{Part 3 : High-Level Tactics}
%$* Self-Discover: Large Language Models Self-Compose Reasoning Structures = 2402.03620
%`Strategy visualisation` (DONE)

\begin{figure*}[ht]
{\scriptsize%
\begin{verbatim}
"Object Segmentation"
    description: "Divide the input grid into distinct objects based on properties like color and connectivity."
    dsl_functions: [as_objects, partition]
"Color Filtering"
    description: "Select objects based on their color."
    dsl_functions: [color_filter]
"Spatial Relationship Analysis"
    description: "Determine the spatial relationship of objects to the grid boundaries or other objects."
    dsl_functions: [bordering, adjacent, manhattan_distance]
"Conditional Selection"
    description: "Choose objects based on a condition (e.g., objects not touching the border)."
    dsl_functions: [keep_if_condition, extract_first_matching]
"Recoloring/Filling"
    description: "Change the color of objects or fill specific areas of the grid with a color."
    dsl_functions: [fill, recolor, paint_onto_grid]
"Grid Transformation"
    description: "Apply transformations to the overall grid such as rotation, mirroring or upscaling."
    dsl_functions: [rot90, rot180, rot270, horizontal_mirror, vertical_mirror, upscale, downscale]
\end{verbatim}%
}
\caption{High-Level Tactics suggested for the sample problem}%
\label{Part3_Tactics}
\end{figure*}

In order to get tactics that might be applicable in a Self-Discover \citep{zhou2024selfdiscoverlargelanguagemodels} framework for ARC, 
the LLM was open-endedly asked to generate `at least 5' high-level tactics, and given a few examples.

The prompt (given in full in Appendix \ref{Appendix_Part3}) extends Part 2:
\begin{itemize}
\item {\bf Create high-level tactics } - this was intentionally very open-ended, 
only specifying that the tactics should be useful if the function solution was not known
\end{itemize}

The tactics suggested by the LLM in Figure \ref{Part3_Tactics} make sense for this example - 
but the real test is whether they are more generally applicable 
(so that a Self-Discover implementation could then `order off the menu' from the available tactics, and then execute them).

\begin{figure}[ht]
\includegraphics[width=0.5\textwidth]{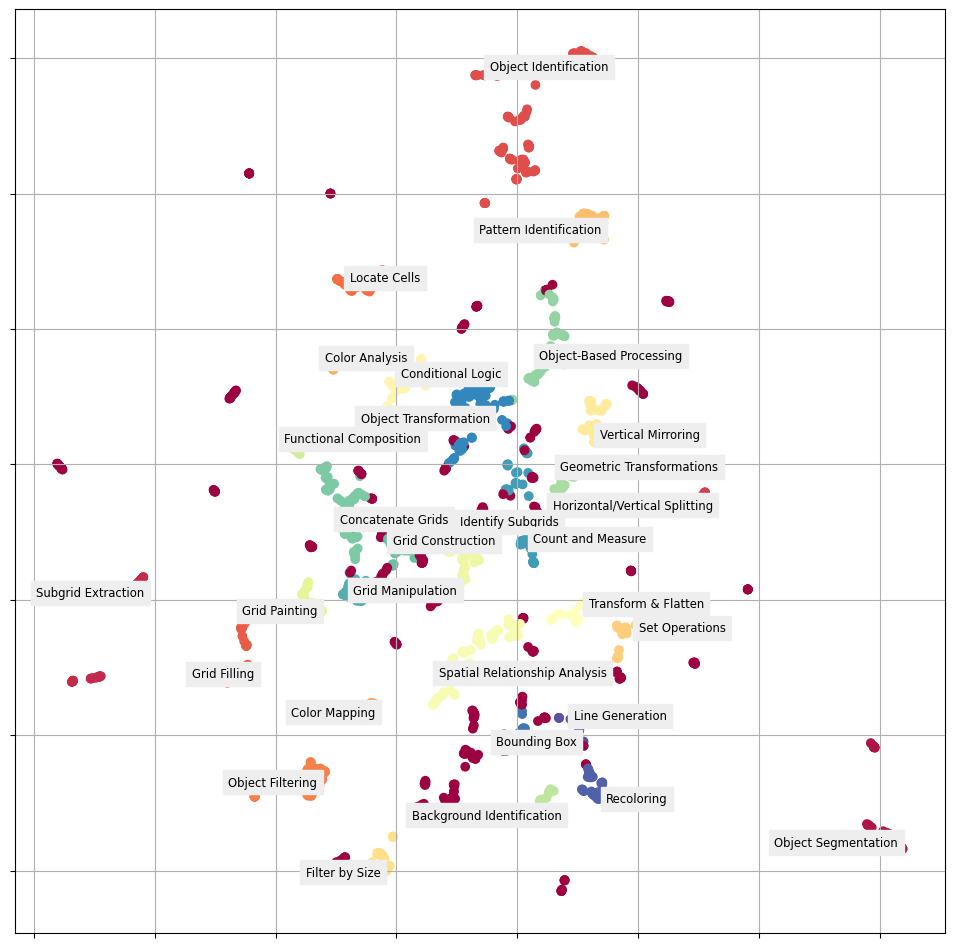}
\caption{Map of Tactics across problems}%
\label{Part3_TacticsMap}
\end{figure}

To investigate this, tactics were gathered from across the dataset outputs,
and then their  sentence-embeddings (provided by {\tt jina-embeddings-v2-base-code}) 
were clustered using UMAP \citep{mcinnes2020umapuniformmanifoldapproximation} and HDBSCAN \citep{Malzer_2020}.
The results are shown graphically in Figure \ref{Part3_TacticsMap}.

The number of points in the top 30 clusters shown suggests that the LLM has, indeed,
been able to surface high-level tactics of the type required for the Self-Discover framework.

%\subsection{Core Knowledge}
%* Core Knowledge
%
%`TODO`: `Core Knowledge histogram` ??
%Core Knowledge per problem somehow...
%Ask ChatGPT about association graphics

\clearpage

\begin{figure*}[ht]
{\scriptsize%
\begin{verbatim}
input: A grid containing multiple colored objects, with one color representing a background and others \
       representing foreground objects.  The foreground objects are generally connected areas \
       of the same color, and are simply connected (i.e. no holes exist within objects).
steps:
  - text: "Identify and separate the objects in the input grid based on their color and connectivity."
    tactic_used: "Object Segmentation"
    core_knowledge: ['Object cohesion']
    variables_input: [I]
    variables_output: [x1]
  - text: "Filter the objects to select only those of a specific color (e.g., black)."
    tactic_used: "Color Filtering"
    core_knowledge: ['Object cohesion']
    variables_input: [x1]
    variables_output: [x2]
  - text: "Identify objects that are not adjacent to the edges/border of the input grid."
    tactic_used: "Spatial Relationship Analysis"
    core_knowledge: ['Object influence via contact', 'Basic Geometry and Topology priors']
    variables_input: [x2, I]
    variables_output: [x5]
  - text: "Recolor/fill the locations of the selected internal objects (those not touching the border) \
           with a new color (e.g., blue)."
    tactic_used: "Recoloring/Filling"
    core_knowledge: ['Object manipulation']
    variables_input: [I, x5]
    variables_output: [O]
output: The output grid is the same as the input grid, except that the internal objects of a specific color \
        are recolored with a new specified color.
\end{verbatim}%
}
\caption{Generic Solution Steps}%
\label{Part4_Steps}
\end{figure*}

\subsection{Part 4 : Solution Steps}

%\noindent
%\begin{minipage}{0.45\textwidth}
In order to obtain a description of the whole task (as if one were a human `describer' for LARC),
the LLM was asked to describe the steps to be taken to transition from the Input Grid to the Output Grid.

The prompt (given in full in Appendix \ref{Appendix_Part4}) extends Part 3:
\begin{itemize}
\item {\bf Description of the input/output grids } - specified to be for the whole task 
\item {\bf Steps required to solve the task } - in `human terms'
\item {\bf Relevant variables names for each step } - so that each of the Parts here can be indexed against each other
\item {\bf Core Knowledge and tactics } - this was so that generic program steps could be provided for fine-tuning other models.  
Note that no specific list of Core Knowledge was supplied, only the rubric given in Appendix \ref{appendix-core-knowledge}
\end{itemize}

The same sample task was used to produce Figure \ref{Part4_Steps}, 
which illustrates a {\it reasonable} ability to describe the process - 
though the Input Grid description includes ``(i.e. no holes exist within objects)'', which is a mistake.

Overall, the LLM tended to be over-cautious in generalising the solving process in this Part
- but this is potentially because of the requirement to specify the variables being used at each step.
%\vspace{100em}
%\end{minipage}

%shown in Table \ref{provability}, where we show results for 2 wordplay generation regimes : 
%Firstly, generate 1 wordplay per question+candidate, but do 5 rollouts for formalisation; 
%Secondly, generate 5 wordplays per question+candidate, for 1 rollout each.
\begin{comment}
\begin{table}[t]
\caption{Frequency of `provability' wins by aggregation method \\
LLM version: 1.0P is Gemini-Pro-1.0, 1.5F is Gemini-Flash-1.5
}
\label{provability}
\vskip 0.15in
\begin{center}
\begin{small}
\begin{sc}
\begin{tabular}{lcccc}
\toprule
Method & LLM  & True & Draw & False \\
       & ver. & Pos  &      & Neg \\
%\midrule
%Completed Proofs & 5x 1 & 27\% & 61\% & 12\% \\
%Fastest Solve    & 5x 1 & 29\% & 55\% & 16\% \\
%Mean solve time  & 5x 1 & 32\% & 52\% & 16\% \\
\midrule
Completed Proofs & 1.0P & 38\% & 59\% &  3\% \\
Fastest Solve    & 1.0P & 38\% & 56\% &  6\% \\
Mean solve time  & 1.0P & 38\% & 56\% &  6\% \\
\midrule
Completed Proofs & 1.5F & 40\% & 55\% &  5\% \\
Fastest Solve    & 1.5F & 40\% & 55\% &  5\% \\
Mean solve time  & 1.5F & 42\% & 53\% &  5\% \\
\bottomrule
\end{tabular}
\end{sc}
\end{small}
\end{center}
\vskip -0.1in
\end{table}
\end{comment}

\clearpage
\begin{comment}
\section{Limitations}
\label{Limitations}

The Prover does not detect a number of potential errors / problems:

\begin{itemize}
\item Cryptic crossword setting `rules' dictate that the clues should contain exactly enough to prove an answer,
      the prover does not check that all valuable words in the clue have been utilised
\item Occasionally, the hint {\tt assert XYZ failed} results in a re-write {\tt assert XYZ==False}, which is cheating
\end{itemize}

With additional effort, the authors believe that these issues are surmountable.

\end{comment}

\section{Conclusions}

%It is increasingly hypothesised that the next-token-prediction task may be
%insufficient to get machines to reason and plan \citep{Kambhampati_2024}.
%By framing the cognitive task of cryptic crossword solving as a reasoning problem 
%that is addressable by LLMs supported by a verification system, 
%this work has sought to bring this reasoning task within the scope of
%what is tractable by \emph{systems} that have components that include LLMs 
%as well as verifiers and coding aids.

The ARC challenge is an important benchmark due to its resistance against brute-force scaling-oriented approaches.
So, while some may argue in favour raising the compute available within the ARC Prize Kaggle environment,
the authors feel that keeping a tight bound on resources will spur more innovation :
Necessity is the mother of invention.

Even though there are strong arguments that `vanilla' LLMs 
will not be capable of learning the abstraction abilities required to tackle the problems head-on from data alone, 
this work illustrates how they might be capable of producing some `Sparks of Abstraction'
which can then be captured for additional processing.

%Clearly, the value in the approach described here (and the code/dataset released) lies in 
%the future use-cases it can be applied to.
%

\subsection{Further Work}

A key goal of this work 
%(as the title suggests) 
has been to find a way to `smuggle the intelligence'
of a commercial LLM into the lower-resource ARC Prize environment.

Preliminary work on the actual ARC Prize challenge has been on-going, 
however the overall system envisioned is still being brought on-line
(while the component parts appear promising, 
integrating them into a cohesive whole is a daunting task).

%Add 'Regular Python' solutions to the input set 
%(the functional style of Hodel's solutions and DSL is more like a CS programming assignment, rather than regular Pythonic code)

%We look forward to competing in the 2025 round of the ARC Prize!
We look forward to making progress on the ARC challenge in the 2025 round of the ARC Prize 
- and would welcome the opportunity for collaboration in the future.

\begin{comment}
\subsection{Citations and References}

Please use APA reference format regardless of your formatter
or word processor. If you rely on the \LaTeX\/ bibliographic
facility, use \texttt{natbib.sty} and \texttt{icml2024.bst}
included in the style-file package to obtain this format.
\end{comment}

\section*{Acknowledgements}

Support for this research was provided by the Google AI/ML Developer Programs team,
including access to the Gemini models and GPUs on Google Cloud Platform. 

%Finally, the authors thank 
%The authors thank 
%for useful comments and 
%the ICML workshop reviewers for their time and valuable feedback.

\begin{comment}
\textbf{Do not} include acknowledgements in the initial version of
the paper submitted for blind review.

If a paper is accepted, the final camera-ready version can (and
usually should) include acknowledgements.  Such acknowledgements
should be placed at the end of the section, in an unnumbered section
that does not count towards the paper page limit. Typically, this will 
include thanks to reviewers who gave useful comments, to colleagues 
who contributed to the ideas, and to funding agencies and corporate 
sponsors that provided financial support.
\end{comment}


\clearpage

%\bibliography{example_paper}
\bibliography{neurips_2024}
\bibliographystyle{icml2024}

%%%%%%%%%%%%%%%%%%%%%%%%%%%%%%%%%%%%%%%%%%%%%%%%%%%%%%%%%%%%%%%%%%%%%%%%%%%%%%%
%%%%%%%%%%%%%%%%%%%%%%%%%%%%%%%%%%%%%%%%%%%%%%%%%%%%%%%%%%%%%%%%%%%%%%%%%%%%%%%
% APPENDIX
%%%%%%%%%%%%%%%%%%%%%%%%%%%%%%%%%%%%%%%%%%%%%%%%%%%%%%%%%%%%%%%%%%%%%%%%%%%%%%%
%%%%%%%%%%%%%%%%%%%%%%%%%%%%%%%%%%%%%%%%%%%%%%%%%%%%%%%%%%%%%%%%%%%%%%%%%%%%%%%
%\newpage
\clearpage
\appendix
\onecolumn

\section{The LLM-legible DSL : {\tt arc-dsl-llm}}
\label{appendix-arc-dsl-llm}

The LLM-legible DSL used in this paper was initially intended to be 
a minor tailoring of {\tt arc-dsl} from \citet{hodel-repo-arc-dsl}.  
However, once a few of the functions had been renamed 
(and consequently more of the DSL code had been read)
the number of changes began to multiply.
Then, the problem of {\tt COLOR} constants being used as integer values, 
violating the typing hints became apparent - 
and the initial minor tailoring became more extensive re-engineering.

The codebase is available at \url{https://github.com/mdda/arc-dsl-llm}, 
and contributions / bugfixes are welcome.  
The code for the dataset that this paper discusses (available at \url{https://github.com/mdda/LLM-abstraction-for-ARC}) 
does not rely on the DSL naming {\it per se}, 
but does currently use the fact that the solutions are all expressed with 
one function call per line (simplifying the parsing / mapping of the code).

\subsection{ Function renaming }

The reasons for changing function names include: 
\begin{itemize}
    \item {\tt dmirror $\rightarrow$ diagonal\_mirror} : Not obvious what the function does without the documentation
    \item {\tt subgrid $\rightarrow$ smallest\_subgrid\_containing} : Similarly...
    \item {\tt product $\rightarrow$ cartesian\_product} : To create a contrast with {\tt multiply}, which behaves as expected
    \item {\tt fork $\rightarrow$ combine\_two\_function\_results} : This may be obvious to a functional programmer, but was confusing to the LLM when reading / generating Python code
    \item {\tt color $\rightarrow$ get\_color} : The LLM sometimes treated {\tt color} as a variable rather than a function
\end{itemize}

For a complete list of the 85 (!) DSL functions that were renamed, please see the repo.
%: \url{https://github.com/mdda/arc-dsl-llm}

\subsection{ {\tt COLOR} constants }

There were several problems with the usage of {\tt COLOR} constants 
in the original {\tt arc-dsl}, which required a large effort to 
correct throughout the codebase (including solutions that made 
assumptions about the numerical values of the defined {\tt COLOR} constants):

\begin{itemize}
    \item Fix assumption that {\tt\verb|COLOR_BLACK==0|} or {\tt\verb|COLOR_BLACK<COLOR_RED|} (for instance)
    \item Add additional constant {\tt\verb|COLOR_BELOW|} (defined to be numerically smaller than other colors) that allows for sort to behave in the way expected by several solutions
    \item Remove usage of {\tt\verb|COLOR_X|} to represent small integers (i.e. non-colors). This was frustrating.
    \item Remove calculation of {\tt \verb|COLOR_X|} values by (for instance) doubling other {\tt\verb|COLOR_Y|} values (!)
\end{itemize}

It is believed that these are largely fixed 
(since in the {\tt 03} notebook the values of the {\tt COLOR} constants is permuted, 
and the validity of the solutions is rechecked).  
However, it is possible that some edge-case were not detected).

\subsection{ Type-hinting }

Although {\tt arc-dsl} appeared to have solid type-hinting, 
it appears that it was not actually checked to be valid.
The {\tt arc-dsl-llm} includes type-hinting 
such that that {\tt\verb|pyright solvers.py|} executes cleanly.
To make it clearer how the solutions worked (including the types), 
two new functions were added : {\tt as\_generic\_tuple} and {\tt make\_cell}.

One problem caused by Python's lack of proper types : 
Integer manipulations of variables that included {\tt COLOR} elements 
(which could occur in, for instance, tuples with `real' integers) 
cannot be `traced through' to ensure type-correctness.
To enable proper `LLM-legible' rendering of both grids and the more complex
interim variables, stricter type-hinting adopted (out of necessity) by the dataset utilities :
{\tt COLOR} constants were remapped to have values in the range {\tt [1009 .. 1019]} 
(corresponding to {\tt [BELOW, BLACK .. PINK]}), 
which was a workable (hacky) solution, given that the ARC Core Knowledge specifies that numerical
quantities used within solutions will be no larger than (say) 20.
Thus, if an integer is $>20$, we know that it can be remapped as a {\tt COLOR} constant 
when rendered for the LLM.

\subsection{ Making {\tt arc-dsl-llm} available as a Python module }

Due to the requirement that {\tt arc-dsl-llm} could be treated as a module 
(to enable {\tt import} for running LLM-generated code that calls DSL functions),
a simple fix was to add a link ``{\tt ./arc-dsl $\rightarrow$ .}'', and adding 
a {\tt \_\_init\_\_.py} to the repo main directory.
This allows the code to run as an imported module 
(without moving the files around - all previous {\tt arc-dsl} change history is preserved).

\subsection{ Confirmation that solutions are still valid }

The {\tt arc-dsl-llm} has been brought up-to-date (as-of 2024-11-12) with the latest
ARC training set fixes, and PRs from the community - 
and all 400 training-set solutions pass cleanly.

Note that, due to how the modularisation fix works,
the command to run the DSL tests and prove the solutions on the test examples
is now : {\tt\verb|python -m arc_dsl.main|}

\begin{comment}
{\scriptsize
\begin{verbatim}
clue:       Chaperone shredded corset (6)
definition: {Chaperone} shredded corset
answer:     ESCORT
wordplay:   (corset)* (*shredded)
\end{verbatim}
}

\end{comment}

% \nopagebreak

%\newpage
\section{Core Knowledge Rubric}
\label{appendix-core-knowledge}

The rubric for the Core Knowledge was reformulated as follows (guided by LLM re-writing) :

{\scriptsize
\begin{verbatim}
## Core Knowledge
Solving ARC problems requires understanding and applying Core Knowledge concepts relating to spatial reasoning, \
object manipulation, and basic mathematical principles. These concepts include:
* **Object cohesion**:
  + Ability to parse grids :
    - identifying distinct objects within the grid based on properties like:
      + For instance: color continuity, spatial contiguity, repeated patterns, or symmetries
    - segmenting the grid into zones or partitions, which can be treated as sub-grids
      + For instance: dividing a grid with delineated quadrants into separate, potentially inter-related sub-grids                   
* **Object persistence**:
  + Objects are assumed to persist despite the presence of noise or occlusion by other objects
    - For example, if a square is partially covered by a triangle, the solver should still recognize \
      the underlying square
    - While generally true, there are cases where objects might disappear or transform significantly
  + In many cases, objects from the input grid persist on the output grid, but in a transformed form but in \
    a transformed form (e.g., rotated, scaled, or recolored)
* **Object influence via contact**: 
  + Many problems feature physical contact between objects
    - For instance: one object being translated until it is in contact with another
    - Other examples: a line extending until it touches another shape; objects snapping to a grid; \
      or an object being 'pushed' by another
* **Basic Geometry and Topology priors**:
  + Geometric and topological reasoning is crucial. Commonly encountered concepts include:
    - Shapes: Lines, rectangles and simple shapes;  Other objects that occur are likely to have simple motifs
    - Transformations: rotation, translation, mirroring, flipping, scaling (overall or horizontal/vertical)
    - Relationships: Containing/contained, inside/outside perimeter, corners, parallel lines, \
      topological connectedness, set relationships (inclusion, intersection, disjointness).
    - Actions: Drawing lines, connecting points, orthogonal projections, copying, repeating objects
    - Self-similarity via symmetries such as rotations and mirroring
* **Numbers and Counting priors**:
  + Many ARC problems involve counting or sorting objects and/or comparing numbers, for instance:
    - Which shape or symbol appears most / least / same number of times?
    - Which object is the largest / smallest?
    - Which objects are the same size / color?
  + Similarly actions being taken might depend on counting and/or comparing numbers
    - For example: Repeating a single shape a number of times depending on the number of different shapes present
  + Simple arithmetic operations (addition, subtraction, multiplication, division), \
    although all quantities featured will be small integers less than (say) 10
* **Goal-directedness prior**:
  + Many ARC problems can be interpreted as depicting a sequence of actions with a specific goal
  + For instance: 
    - A problem might combines the concepts of "line extrapolation", "turning upon hitting an obstacle", \
      and "efficiently reaching a goal"
    - Arranging objects to fill a container or constructing a symmetrical pattern
  + Some ARC problems might imply a need for planning or simulating steps towards a solution
* **Compositionality**:
  + Successfully solving ARC problems often requires chaining the above concepts together
    - For instance: First identifying simply connected components (cohesion), then counting them (numerical), \
      and finally replicating the largest component multiple times side-by-side (geometry)
    - For instance: First grouping shapes by color (cohesion and color), sorting them by size (numerical), \
      recoloring the most frequent (numerical and color), and reflecting it across \
      a vertical axis (geometry and symmetry)
\end{verbatim}
}

\newpage
\section{Gemini Prompting Scheme}
\label{appendix-gemini-prompt}

The overall rubric for the ARC challenge task follows :

{\scriptsize
\begin{verbatim}
# ARC Challenge problems
Each problem in the ARC Challenge requires understanding the way in which several "input grids" \
can be transformed into corresponding "output grids".  
Several demonstration pairs are shown, and the solution involves describing how an unknown "output grid"  \
can be derived from the given test "input grid".
To do this, we will be doing extensive code analysis.
\end{verbatim}
}

\subsection{Part 1 - Code Commenting Prompts}
\label{Appendix_Part1}

{\scriptsize
\begin{verbatim}
### Part 1 : Add comments to original solution
Add comments into the program code for function `solver_virtual(I)` above, at the points indicated by `# comment`.
If it makes sense, comments can be skipped, so that lines of code are combined into more reasonable code blocks.
Each code block can be as short as one line, or as long as necessary to encompass a complete subtask.
Each set of comments should relate to the code block that follows.

#### Part 1 Answer Format
Your answer should repeat the program code of `solver_virtual(I)` above, with the comments included according \
to the code blocks you decide.  
Each set of comments should be in the following format:
* `# Input: ` What input the code is expecting at that point (in terms of types, and in terms \
              of the overall goal of the solution)
* `# Goal: ` What the goal of the next line of code are (both locally, and how it relates \
             to the overall goal of the solution).  
* `# Output: ` What the expected output of this block (in terms of types, and in terms \
             of the overall goal of the solution)
* (optional) `# Core Knowledge: ` If any elements of Core Knowledge are relevant to the block, \
                                  describe them in an additional comment line.
\end{verbatim}
}

\subsection{Part 2 - Code Refactoring Prompts}
\label{Appendix_Part2}

{\scriptsize
\begin{verbatim}
### Part 2 : Create reusable components
Create a new version of `solver_virtual(I)` from Part 1 called `solver_virtual_chunked(I)` \
which has the same functionality.
To create `solver_virtual_chunked(I)`, examine each line of code (and surrounding lines):
* move natural blocks of code (consisting of several lines of code each) into separate new functions, \
  with a call from `solver_virtual_chunked(I)`.
* blocks of code must return concrete variables. 
* Callables should only be used be within a block
* if there are lines that are not easily isolated, leave them unchanged in `solver_virtual_chunked(I)`.
Comments in the same format as Part 1 should be added to each line of `solver_virtual_chunked(I)`.

#### Part 2 Answer Format
The following example illustrates the format of two function components and `solver_virtual_refactored(I)`:
```python
def recolor_single_cell_objects(pairs: FrozenSet, color: Color) -> FrozenSet:  # New function, which calls \
                                                                                 at least 2 DSL functions
  # Input: pairs (FrozenSet), color (Color), pairs of single-cell and grey objects
  # Goal: Recolor each single-cell object based on its adjacent object's color.
  # Output: recolored_objects (FrozenSet), a set of locations and recolored single-cell objects.
  # Core Knowledge: Object transformation (recoloring), Compositionality
  recoloring_function = combine_two_function_results(recolor, compose(color, get_first), get_last) \
                           # variables named appropriately
  recolored_objects = transform_and_flatten(recoloring_function, pairs) # variables named appropriately
  return recolored_objects

# ...  other new functions here

def solver_virtual_chunked(I):  # This function calls the new functions, replacing suitable chunks. \
                                  Variable names in this function are the same as in `solver_virtual`
  # Input: I (Grid), the input grid.
  # Goal: Identify and separate objects within the input grid.
  # Output: x1 (Objects), a set of objects identified in the input grid.
  # Core Knowledge: Object cohesion (parsing grids, identifying distinct objects based on spatial contiguity)
  x1 = as_objects(I)  # Retain original code (and variable names) if not moved to new function

  # Input: x1 (Objects), a set of objects.
  # Goal: Filter objects based on their size (select only single-cell objects).
  # Output: x2 (Objects), a subset of x1 containing only single-cell objects.
  # Core Knowledge: Numbers and Counting priors (size filtering).
  x2 = size_filter(x1, 1)   # Retain original code (and variable names) if not moved to new function

  # ... other lines here - with each block also having comments in the format of Part 1.

  # Input: x2 (FrozenSet), pairs of single-cell and objects
  # Goal: Recolor each single-cell object based on its adjacent object's color.
  # Output: x9 (FrozenSet), a set of locations and recolored single-cell objects.
  # Core Knowledge: Object transformation (recoloring), Compositionality
  x9 = recolor_single_cell_objects(x2, GREY)  # Call new function, retain original variable names in caller 

  # Input: I (Grid), input grid; x9 (FrozenSet), recoloring instructions.
  # Goal: Update input grid by painting the recolored objects onto it.
  # Output: O (Grid), the output grid after recoloring.
  # Core Knowledge: Object manipulation (painting), Compositionality.
  O = paint_onto_grid(I, x9)   # Retain original code if not refactored
  return O
```
\end{verbatim}
}

\subsection{Part 3 - High-Level Tactics Prompts}
\label{Appendix_Part3}

{\scriptsize
\begin{verbatim}
### Part 3 : High-level tactics
Outline potential high-level tactics that could be used to solve this problem, \
if `solver_virtual(I)` was unknown.

#### Part 3 Answer Format
Fill in the following YAML structure (the comments explain the intent of the entries):
```yaml
tactics:
  - heading: "" # A short name for the tactic 
    description: "" # A description of the tactic 
    dsl_functions: [] # A list of relevant DSL functions (as appropriate)
```
Return 5 or more tactics in this format.

#### Part 3 Examples
Some examples of tactics: 
```yaml
tactics:
  - heading: "Better Representation"
    description: "Seek a better representation of the input/output grid"
    dsl_functions: [as_objects]
  - heading: "Filter by Property"
    description: "From the list, select according to a property"
    dsl_functions: [size_filter, most_common_color, extract_first_matching, equals]
  - heading: "Combine Results"
    description: "Combine previous results into final grid"
    dsl_functions: [fill, paint_onto_grid]
```
\end{verbatim}
}

\subsection{Part 4 - Overall Solution Prompts}
\label{Appendix_Part4}

{\scriptsize
\begin{verbatim}
### Part 4 : Overall solution description
Describe the high-level steps involved in solving the overall Problem.  
This requires stating the overall expected contents of the Input grid, a sequence of steps required \
to solve the problem, and the expected contents of the Output grid.
The sequence of steps should be expressed in human form (not necessarily corresponding directly to lines of code). 
The steps should be described generically (i.e. don't use specific color names or shape descriptions) \
so that the steps could be reused for other problems.

#### Part 4 Answer Format
Fill in the following YAML structure (the comments explain the intent of the entries):
```yaml
input: "" # What input should be expected for the problem
steps: # An array with elements that correspond to each high-level step
  - text: ""  #  describes this key part of solving the problem
    tactic_used: ""  # the tactic heading from Part 3 that is most relevant to this step
    core_knowledge: []  # if any elements of Core Knowledge are relevant to this step, list them \
                          (eg: ['Object Manipulation', ...])
    variables_input: [] # if any variables in Part 1 are needed before doing this step, list them (eg: [x3, x4])
    variables_output: [] # if any variables in Part 1 are created by this step, list them (eg: [x3, x4])
output: "" # What output should be expected for the problem solution
```
\end{verbatim}
}

\end{document}